\documentclass[11pt]{article}

\usepackage[final]{acl}
\usepackage{array} 

\usepackage{microtype}
\usepackage{hyperref}

\usepackage[english,bidi=default]{babel} 
\babelfont{rm}{TeXGyreTermesX} 
\babelprovide[import]{hindi}
\babelfont[*devanagari]{rm}{Lohit Devanagari}
\babelprovide[import]{arabic}
\babelfont[*arabic]{rm}{Noto Sans Arabic}

\newcommand\figref[1]{Figure~\ref{fig:#1}}
\newcommand\tabref[1]{Table~\ref{tab:#1}}

\usepackage{tabularx}
\usepackage{enumitem}
\usepackage{booktabs}  

\usepackage{listings}

\usepackage[most]{tcolorbox}
\tcbuselibrary{listings,breakable}
\tcbset{
  colback=gray!5!white,
  colframe=black!30,
  boxrule=.4pt,
  arc=1mm
}   

\newcommand\pbench[0]{PBIG-DATA}
\newcommand\krip[0]{Krippendorff's $\alpha$}

\newcommand\llmjudge[0]{LLM-as-a-Judge}

\title{Aggregate vs. Personalized Judges in Business Idea Evaluation: \\ Evidence from Expert Disagreement}

\author{
  \textbf{Wataru Hirota}$^{1}$, \textbf{Tomoki Taniguchi}$^{2}$, \textbf{Tomoko Ohkuma}$^{2}$, \textbf{Kosuke Takahashi}$^{1}$, \\
  \textbf{Takahiro Omi}$^{1}$, \textbf{Kosuke Arima}$^{1}$, \textbf{Takuto Asakura}$^{1}$, \textbf{Chung-Chi Chen}$^{3,4}$\footnotemark[1], \textbf{Tatsuya Ishigaki}$^{4}\thanks{The last two authors are co-leads of this project.}$ \\
  $^{1}$Stockmark Inc $^{2}$Asahi Kasei Corporation
  $^{3}$ National Institute of Informatics \\
  $^{4}$National Institute of Advanced Industrial Science and Technology\\
  \texttt{wataru.hirota@stockmark.co.jp chen@nii.ac.jp ishigaki.tatsuya@aist.go.jp}}

\begin{document}
\maketitle
\begin{abstract}
Evaluating LLM-generated business ideas is often harder to scale than generating them.
Unlike standard NLP benchmarks, business idea evaluation relies on multi-dimensional criteria such as feasibility, novelty, differentiation, user need, and market size, and expert judgments often disagree.
This paper studies a methodological question raised by such disagreement: should an automatic judge approximate an aggregate consensus, or model evaluators individually?
We introduce \pbench{}, a dataset of approximately 3,000 individual scores across 300 patent-grounded product ideas, provided by domain experts on six business-oriented dimensions:
specificity, technical validity, innovativeness, competitive advantage, need validity, and market size.
Analyses show substantial expert disagreement on fine-grained ordinal scores, while agreement is higher under coarse selection, suggesting structured heterogeneity rather than random noise.
We then compare three judge configurations: a rubric-only zero-shot judge, an aggregate judge conditioned on mixed evaluator histories, and a personalized judge conditioned on the target evaluator's scoring history.
Across dimensions and model sizes, personalized judges align more closely with the corresponding evaluator than aggregate judges, and evaluator agreement correlates with similarity of judge-generated reasoning only under personalized conditioning.
These results indicate that pooled labels can be a fragile target in pluralistic evaluation settings and motivate evaluator-conditioned judge designs for business idea assessment.
\end{abstract}

\section{Introduction}

Large language models (LLMs) make it easy to generate large numbers of product ideas.
The practical bottleneck is not generation but evaluation.
Organizations must decide which ideas are sufficiently concrete, technically feasible, differentiated, and commercially meaningful to justify further investment.
These decisions are often made by multiple experts with different backgrounds, such as technical reviewers and business strategists.

Unlike tasks where evaluation targets factual correctness or task completion, business idea evaluation is multi-dimensional and judgment-driven.
Even when reviewers follow the same rubric, they may apply different assumptions about feasibility, novelty, risk, or market opportunity.
In practice, this leads to persistent disagreement across experts.
Such disagreement is not necessarily annotation noise; it reflects heterogeneous standards about what constitutes a promising idea.

Most \llmjudge{} approaches assume that a single scoring standard exists.
Under this assumption, labels from multiple reviewers are aggregated, and the judge is optimized to reproduce this pooled signal.
However, when expert disagreement is systematic, aggregation may obscure meaningful differences between evaluators.
This raises a methodological question for evaluation design:
should an automatic judge approximate an aggregate consensus, or should it model evaluators individually?

This paper studies this question in the context of business-oriented product idea evaluation.
We introduce \pbench{}, a dataset of approximately 3,000 individual scores across 300 patent-grounded product ideas, provided by domain experts along six business-oriented dimensions:
specificity, technical validity, innovativeness, competitive advantage, need validity, and market size.

Our empirical findings reveal a substantial gap between common evaluation assumptions and actual expert behavior.
First, inter-annotator agreement on fine-grained ordinal scores is often close to zero and occasionally negative, indicating that expert scoring does not converge to a single shared scale.
At the same time, agreement increases when the task is framed as coarse selection, suggesting that disagreement reflects heterogeneous but structured standards rather than random noise.
Second, this heterogeneity has direct implications for \llmjudge{} design.
We compare three judge configurations: a rubric-only zero-shot judge, an aggregate judge conditioned on mixed evaluator histories, and a personalized judge conditioned on the target evaluator's own scoring history.
Across dimensions, personalized judges align more closely with the corresponding evaluator than aggregate judges.
This indicates that individual evaluators are internally consistent even when they disagree with each other.

Taken together, these results suggest that business idea evaluation is inherently pluralistic.
Treating pooled labels as a single ground truth may be an inadequate target for automatic judging.
Modeling evaluator-specific standards provides a more faithful representation of expert judgment, but it also highlights how far current judge designs are from fully supporting heterogeneous decision processes in real-world settings.

This paper makes the following contributions:
\begin{enumerate}
    \item A dataset of expert-scored product ideas under business-oriented criteria.
    \item A quantitative analysis of structured expert disagreement.
    \item Evidence that aggregate and personalized judge designs behave differently under heterogeneous standards, with implications for evaluation methodology in practical ideation systems.
\end{enumerate}

\section{Related Work}

\subsection{Human Evaluation in Creative Generation}

Human evaluation plays a central role in creative natural language generation tasks.
Prior surveys have shown that inter-annotator agreement is often low in creative settings, especially when tasks involve open-ended judgments such as novelty and emotional impact~\citep{hamalainen-alnajjar-2021-human}.
\citet{amidei-etal-2019-agreement} argue that low agreement does not necessarily indicate unreliable data, but may reflect irreducible variability in human language interpretation.
They recommend complementing agreement metrics with correlation-based analyses to better understand evaluation reliability.

Recent large-scale studies further confirm that even domain experts disagree substantially when evaluating generative outputs.
For example, \citet{Si2025Can} report significant variation among NLP researchers assessing the novelty of LLM-generated research ideas.
These findings suggest that disagreement is a structural property of creative evaluation rather than annotation noise.
Our work extends this line of investigation to business-oriented product idea evaluation, focusing specifically on how disagreement affects the design of automatic judges.

\subsection{LLMs for Ideation}

LLMs have increasingly been used to support ideation tasks across domains.
Interactive systems such as Wordcraft demonstrate how language models can assist human writers in creative writing workflows \citep{yuan-etal-2022-wordcraft}.
In scientific and research ideation, multi-agent LLM frameworks have been explored to improve diversity and feasibility of generated proposals \citep{ueda-etal-2025-exploring}.
Large-scale evaluations of idea generation benchmarks also examine LLM performance on novelty and feasibility metrics \citep{Si2025Can}.

While these studies focus primarily on improving generation quality, evaluation methodology remains comparatively underexplored.
Most ideation benchmarks rely either on automatic scoring or on pooled human labels treated as ground truth.
In contrast, our work centers on the evaluation stage itself and investigates how heterogeneous expert standards influence judge modeling.

\subsection{LLM-as-a-Judge and Evaluation Modeling}

LLMs are increasingly used as automatic judges for diverse NLP tasks.
Early evidence shows that large models can approximate human preferences under certain prompting strategies \citep{openai2023gpt4}.
Subsequent studies systematically examine judge robustness and vulnerabilities, highlighting issues such as prompt sensitivity and alignment bias across domains \citep{thakur2025judging,tan2025judgebench}.
Self-preference bias further complicates judge reliability when models evaluate outputs similar to their own generations \citep{wataoka2025selfpreferencebiasllmasajudge}.

Recent work explores whether LLMs can model individual evaluators rather than aggregate labels.
\citet{dong-etal-2024-llm} investigate personalized judging and show that conditioning on evaluator-specific history can improve alignment.
However, these studies primarily focus on general NLP evaluation tasks.
The implications of personalization under systematically low inter-annotator agreement remain insufficiently studied.

Our work contributes to this discussion by examining aggregate versus personalized judge configurations in a business ideation setting.
By explicitly quantifying expert disagreement and comparing judge alignment under heterogeneous standards, we provide evidence that evaluation modeling choices must account for pluralistic expert judgments rather than assuming a single unified ground truth.

\section{Data and Evaluation Setup}
\label{sec:data_setup}

This section describes the data, scoring dimensions, and annotation protocol used in our study.
The objective is to define an evaluation setting where multiple experts score the same type of business idea under a shared rubric, while allowing for heterogeneous standards and incomplete consensus.

\subsection{Scoring Dimensions and Rubric}
\label{sec:rubric}

Each idea is scored along six business-oriented dimensions: specificity, technical validity, innovativeness, competitive advantage, need validity, and market size, summarized in \tabref{dimensions_overview} (the full level-by-level rubric is provided in Appendix~\ref{sec:rubric_full}).
Our six dimensions are synthesized from two widely used frameworks for early-stage product and innovation assessment applied to LLM-generated patent-grounded ideas.
The NABC framework~\citep{carlson2006innovation} frames new product assessment around Need, Approach, Benefit, and Competition, emphasizing that promising ideas should combine a concrete user need, a credible technical approach, differentiated benefits, and awareness of the competitive landscape.
Cooper's Stage-Gate model~\citep{cooper1990stagegate} operationalizes early-stage screening along feasibility, differentiation, and market attractiveness criteria.

The scale for each dimension is chosen to match its natural granularity in screening decisions rather than to enforce a uniform range.
Specificity, technical validity, and competitive advantage use 1--4 scales because they admit natural four-level gradations (from ``unusable'' to ``production-ready'' in the case of technical validity, for example).
Innovativeness uses a 1--5 scale to give evaluators an additional level to distinguish ``surprising but not groundbreaking'' from ``clearly innovative''.
Need validity and market size use 0--3 scales because the lowest level encodes a categorical exclusion (``not a B2B product'') rather than a low-quality gradation; collapsing this into a 1-based scale would conflate ``non-applicable'' with ``applicable but weak''.

\begin{table}[t]
\centering
\small
\renewcommand{\arraystretch}{0.95}
\setlength{\tabcolsep}{4pt}
\begin{tabular}{lccl}
\toprule
Dimension & Scale & Threshold & Focus \\
\midrule
Specificity & 1--4 & $> 2$ & Clarity of idea \\
Technical validity & 1--4 & $> 1$ & Feasibility \\
Innovativeness & 1--5 & -- & Novelty \\
Competitive advantage & 1--4 & -- & Differentiation \\
Need validity & 0--3 & -- & User need \\
Market size & 0--3 & -- & Adoption scale \\
\bottomrule
\end{tabular}
\caption{
Business-oriented scoring dimensions.
The \emph{Threshold} column gives the score cutoff that must be exceeded for downstream dimensions to be scored under the staged screening protocol. ``--'' indicates a dimension that is only gated by upstream thresholds.
Full rubric is provided in Appendix~\ref{sec:rubric_full}.}
\label{tab:dimensions_overview}
\end{table}

\subsection{Product Ideas and Patents}
\label{sec:ideas}

\pbench{} contains approximately 300 product ideas generated by LLM-based systems.
Each idea is grounded in a patent document and consists of four fields: product title, product description, implementation, and differentiation.

The ideas were produced by multiple independent ideation systems with different prompting and agentic designs, including divergent and convergent ideation, multi-agent discussion, and iterative rewriting pipelines
\citep{yoshiyasu-2025-team,kanumolu-etal-2025-agent,xu-etal-2025-mk2,terao-tachioka-2025-collaborative,hoshino-etal-2025-business,shimanuki-etal-2025-business}; see Appendix~\ref{sec:gen_overview} for an overview.
Using outputs from multiple systems reduces dependence on any single generation strategy and helps ensure that disagreement patterns are not artifacts of a specific approach.

The input patents are sampled from the USPTO corpus and span three technical areas: natural language processing (NLP), computer science (CS), and material chemistry (MatChem).
Domain diversity is important because feasibility and market judgments may depend on background knowledge and domain-specific assumptions.

\subsection{Expert Annotation Protocol}
\label{sec:annotation}

We employ domain experts with both technical and business backgrounds.
Across all three domains, we require at least five years of professional experience in the respective domain.
Technical experts are additionally required to have a record of publishing peer-reviewed research papers in the domain, while business experts are required to have experience in either consulting or B2B new-business development within the domain.
Experts are assigned to domains where they have relevant experience and evaluate only patents they can confidently interpret.

The annotation process follows a staged screening protocol to avoid forcing scores when a dimension cannot be meaningfully assessed.
All experts first score specificity.
If an idea does not meet a minimum specificity threshold (see the \emph{Threshold} column of \tabref{dimensions_overview}), downstream dimensions are not scored.
On the technical side, technical validity is scored only when specificity passes the threshold, and innovativeness and competitive advantage are scored only when both specificity and technical validity pass their thresholds.
On the business side, need validity and market size are scored only when specificity passes the threshold.
This protocol produces missing scores by design.
We treat missingness as part of the evaluation process rather than annotation noise.
The staged screening protocol is illustrated in Appendix~\ref{sec:screening_protocol}.

\tabref{data_stats} summarizes the subset of patents and ideas selected for expert scoring in each domain, and the resulting number of score entries after staged screening.
\begin{table}[t]
\centering
\small
\renewcommand{\arraystretch}{0.95}
\setlength{\tabcolsep}{4pt}
\begin{tabular}{lccrr}
\toprule
Domain & Experts & Patents & Ideas & Annotations \\
\midrule
NLP & 12 & 46 & 100 & 1,055 \\
CS & 11 & 48 & 97 & 984 \\
MatChem & 4 & 22 & 110 & 1,070 \\
\midrule
Total & -- & 116 & 307 & 3,109 \\
\bottomrule
\end{tabular}
\caption{
Dataset coverage by domain.
Annotations count idea–dimension score entries after staged screening.
}
\label{tab:data_stats}
\end{table}

\subsection{Problem Formulation for Automatic Judges}
\label{sec:judge_formulation}

Given a patent and a product idea, an automatic judge predicts a score for a target dimension.
We compare three judge configurations that correspond to different assumptions about the target signal under heterogeneous expert standards:

\begin{itemize}
    \item Zero-shot judge: the model predicts scores using only the rubric and task instructions, without access to prior human scoring examples.
    \item Aggregate judge: the model is conditioned on scoring histories sampled from multiple evaluators, treating pooled behavior as the target.
    \item Personalized judge: the model is conditioned on scoring history from the target evaluator, aiming to reproduce evaluator-specific standards rather than a pooled consensus.
\end{itemize}

These configurations allow us to test whether aggregation is an appropriate modeling target when expert disagreement is systematic.

\begin{figure}[t]
    \centering
    \includegraphics[width=0.85\columnwidth]{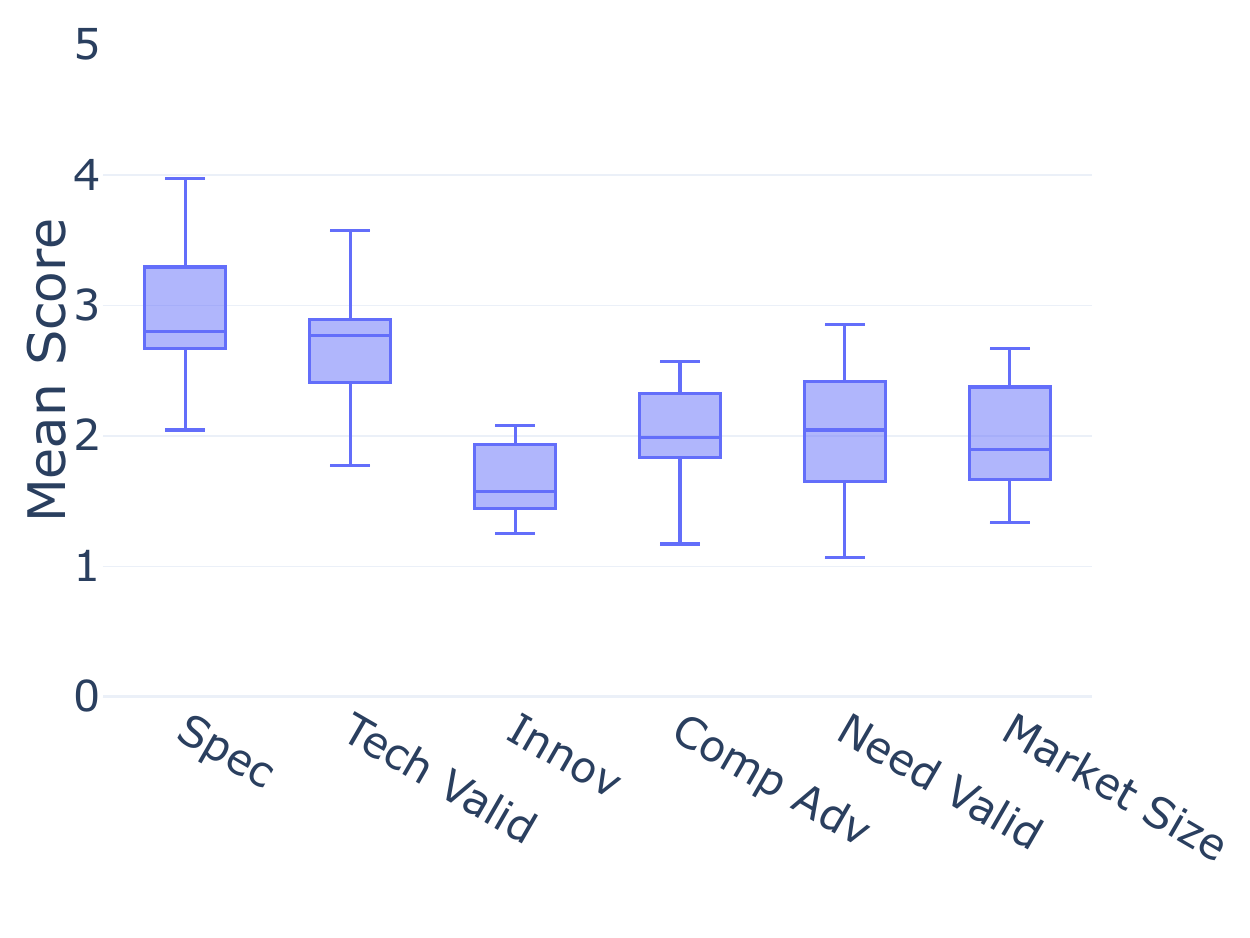}
    \caption{Distribution of per-evaluator mean scores for each dimension.}
    \label{fig:score_dist}
\end{figure}

\section{Expert Disagreement in Business Idea Scoring}
\label{sec:disagreement}

This section characterizes expert disagreement in \pbench{}.
The goal is to understand what the expert labels represent as a target for automatic judging.
If scores do not converge to a shared ordinal scale, then treating pooled labels as a single ground truth can be a misspecified objective.

\subsection{Variation in Evaluator Scoring Scales}
\label{sec:score_dist}

We first examine differences in scoring strictness across evaluators.
For each evaluator and each dimension, we compute the mean score over the items that the evaluator rated.
\figref{score_dist} summarizes these per-evaluator means.

The distributions show substantial inter-evaluator variation for many dimensions.
Variation is especially pronounced for need validity and market size, where judgments depend on implicit assumptions about target users and adoption context.
This pattern suggests that the same score can reflect different thresholds across evaluators, making direct aggregation of ordinal labels problematic.

\subsection{Fine-Grained vs. Coarse Agreement}
\label{sec:agreement_metrics}

We quantify agreement at two levels that correspond to different practical uses of scores.

Fine-grained agreement measures whether evaluators agree on ordinal scores.
We compute \krip{}~\citep{krippendorff2011computing} for each dimension and domain.
Coarse agreement measures whether evaluators select similar subsets of strong ideas even if they disagree on exact scores.
For each evaluator and dimension, we define the above-median items (median computed within that evaluator) and compute the mean Jaccard similarity between evaluator pairs that share at least 10 overlapping scored items.

\tabref{expert_disagreement} reports both metrics.
Fine-grained agreement is often close to zero, and in some cases negative, indicating that evaluators do not rank ideas on a shared scale.
Coarse agreement is higher across many settings, suggesting that evaluators share some common structure in identifying stronger ideas even when they disagree on exact scores.

\begin{table}[t]
\centering
\small
\renewcommand{\arraystretch}{0.95}
\setlength{\tabcolsep}{3pt}
\begin{tabular}{lrrrrrr}
\toprule
& \multicolumn{3}{c}{Fine ($\alpha$)}
& \multicolumn{3}{c}{Coarse (Jac.)} \\
\cmidrule(lr){2-4} \cmidrule(lr){5-7}
Dim. & NLP & CS & Mat. & NLP & CS & Mat. \\
\midrule
Spec. & 0.06 & -0.11 & 0.04 & 0.45 & 0.48 & 0.45 \\
Tech. val. & -0.03 & -0.40 & -0.28 & 0.50 & 0.33 & 0.42 \\
Innov. & 0.33 & 0.47 & 0.46 & 0.71 & 0.43 & 0.54 \\
Comp. adv. & -0.08 & 0.24 & -0.02 & 0.71 & 0.33 & 0.46 \\
Need & -0.23 & 0.02 & 0.05 & -- & -- & 0.89 \\
Market & 0.48 & -0.31 & 0.08 & -- & -- & 0.57 \\
\bottomrule
\end{tabular}
\caption{
Expert disagreement summary.
Fine-grained agreement is Krippendorff's $\alpha$.
Coarse agreement is mean Jaccard similarity of above-median sets.
}
\label{tab:expert_disagreement}
\end{table}

\begin{figure*}[t]
    \centering
    \includegraphics[width=\textwidth]{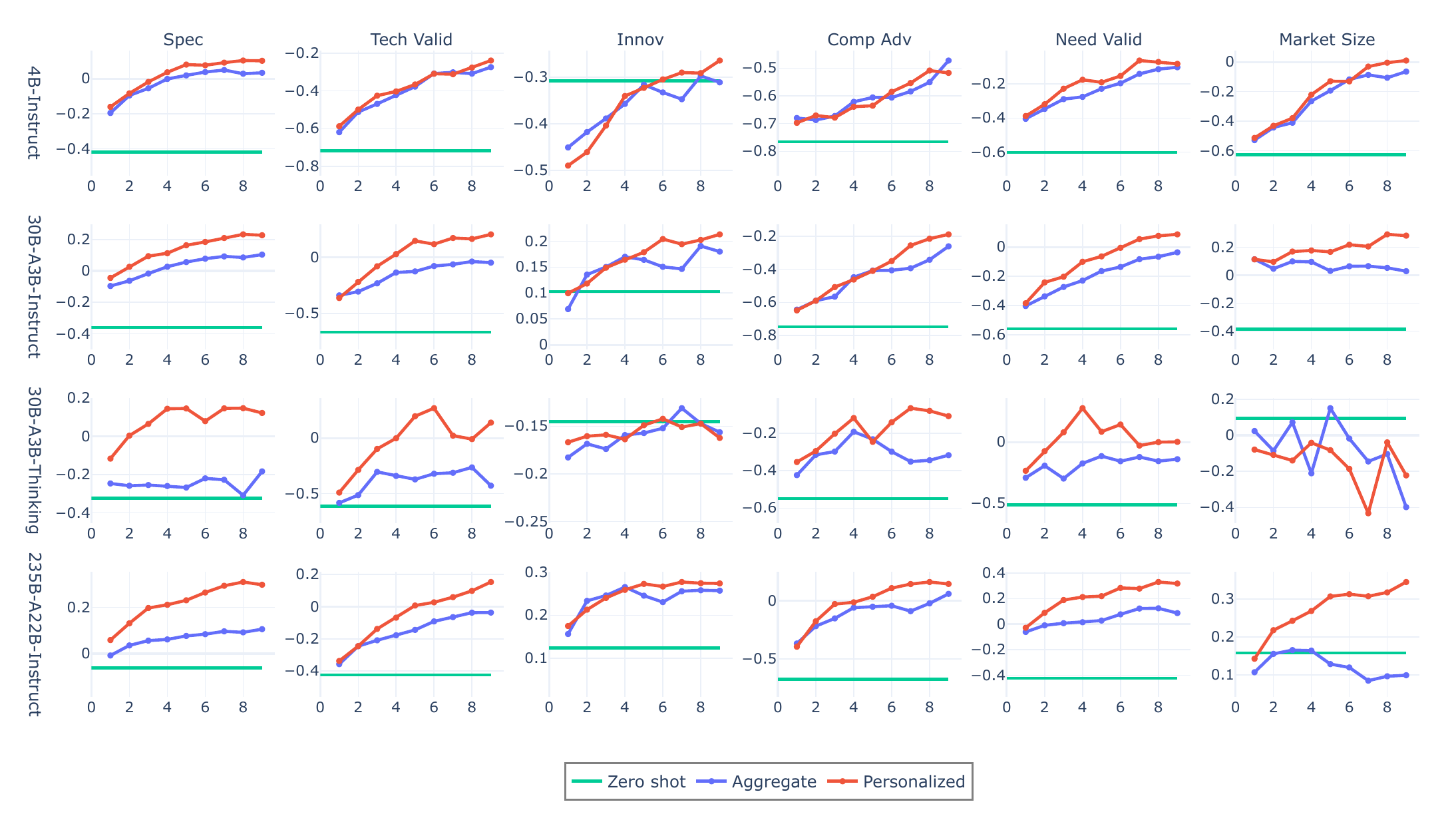}
    \caption{
    Alignment between automatic judges and expert annotations,
    measured by Krippendorff's $\alpha$. X-axis is the number of few-shot examples.
    }
    \label{fig:judge_alignment}
\end{figure*}

These results support two conclusions that matter for judge design.
First, disagreement is systematic rather than purely noise, and pooled ordinal labels should not be assumed to represent a stable ground truth.
Second, because evaluators can be consistent in coarse selection while differing in score calibration, judge alignment should be evaluated with attention to which target signal is being modeled.

\subsection{Implications for Judge Modeling}
\label{sec:disagreement_implications}

The disagreement patterns in \tabref{expert_disagreement} motivate the aggregate versus personalized comparison in the next section.
If the correct target is a shared scoring standard, conditioning on mixed evaluator histories should be sufficient.
If evaluators apply heterogeneous but internally consistent standards, evaluator-specific conditioning may better reproduce expert behavior.
We next test these alternatives using zero-shot, aggregate, and personalized judges.

\section{Aggregate vs. Personalized Judges}
\label{sec:judges}

Section~\ref{sec:disagreement} showed that expert scores do not collapse into a shared ordinal scale.
This section evaluates how different judge configurations behave under such heterogeneous standards.

\subsection{Experimental Setup}

We evaluate the zero-shot, aggregate, and personalized judges defined in Section~\ref{sec:judge_formulation}.
The goal is to measure how closely each configuration aligns with expert annotations.
We use four Qwen3 models of varying sizes \citep{qwen3technicalreport}:
4B-Instruct-2507,
30B-A3B-Instruct-2507,
30B-A3B-Thinking-2507,
and 235B-A22B-Instruct-2507.
The 30B-A3B-Thinking model includes reasoning steps.
Results for GPT-5 mini are reported in Appendix~\ref{sec:gpt5_results}.
Section~\ref{sec:coarse_metrics} complements the Krippendorff's $\alpha$ results with coarse selection metrics.

Few-shot examples are drawn from the pool of scored instances in the same domain and same dimension, but always from ideas grounded in different patents than the target to prevent leakage.
The personalized judge and the aggregate judge use the same selection logic except for one constraint: for the personalized judge, examples must be scored by the same evaluator as the target; for the aggregate judge, examples must be scored by a different evaluator, creating a cross-evaluator conditioning set.
The target item itself is always excluded from the conditioning set.
For each shot count, examples are sampled without replacement with three random-seed variants.
We note that the evaluation follows a leave-one-out protocol rather than a traditional train/test split.
Each target instance is scored by a judge conditioned on examples drawn from the remaining instances in the same domain and dimension but from different patents, so no patent appears simultaneously in the target and in its own conditioning set.

The judge predicts a Likert score together with a self-reported confidence value in the range 0--100 (see Appendix~\ref{sec:prompt_schema} for the prompt).
Following \citet{dong-etal-2024-llm}, we discard predictions with confidence below 80. Appendix~\ref{sec:discard_rates} shows the discard rates.

Three runs with different random seeds are performed,
and majority voting determines the final prediction.
Agreement between predictions and expert annotations
is measured using Krippendorff's $\alpha$,
consistent with the fine-grained metric in Section~\ref{sec:agreement_metrics}.

\subsection{Alignment with Expert Annotations}
\figref{judge_alignment}
shows alignment across models and dimensions.
For most dimensions and model sizes,
personalized judges agree more closely with the corresponding evaluator than aggregate judges.
The gap becomes more pronounced for larger models.

Aggregate judges typically outperform zero-shot judges,
indicating that historical examples improve calibration.
However, aggregate conditioning consistently underperforms personalized conditioning.
This suggests that pooled scoring histories encode an averaged behavior
that does not fully represent any individual evaluator's standard.

Zero-shot performance remains near zero in many dimensions,
demonstrating that the rubric alone does not define a unique scoring scale.
Historical examples are necessary for calibration,
but the type of history matters.
Appendix~\ref{sec:case_study} presents a qualitative case study in which aggregate and personalized judges assign different innovativeness scores to the same idea, illustrating how evaluator-specific calibration manifests in individual predictions.

\begin{figure}[t]
    \centering
    \includegraphics[width=0.85\columnwidth]{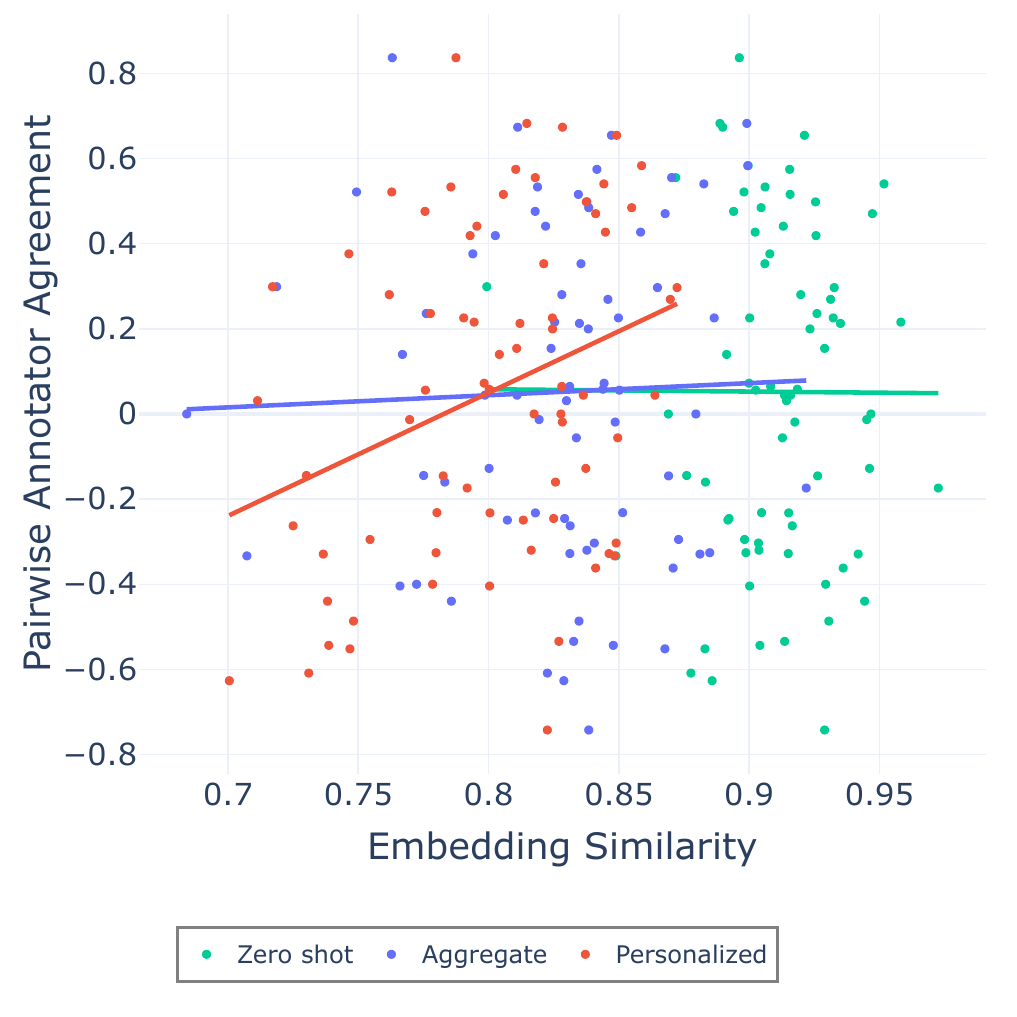}
    \caption{
    Relationship between evaluator agreement and similarity of judge-generated reasoning texts.
    A positive trend appears only under personalized conditioning.
    }
    \label{fig:reasoning_similarity}
\end{figure}

\subsection{Coarse Selection Metrics}
\label{sec:coarse_metrics}

Section~\ref{sec:agreement_metrics} argued that expert disagreement is more pronounced at the fine-grained ordinal level than at the coarse selection level.
To check whether the personalization advantage also extends to coarse selection behavior, we report two additional metrics computed on the 9-shot judges (the maximum shot count in our experiments).

\paragraph{Above-median Jaccard similarity.}
For each evaluator and dimension, we compute the Jaccard similarity between the evaluator's and the judge's above-median item sets (\tabref{coarse_jaccard}).
Personalized judges achieve higher Jaccard similarity than aggregate judges on five of the six dimensions.

\paragraph{Top-50\% overlap.}
We also compute the fraction of each expert's top-50\% items that the judge also ranks in its own top-50\% (\tabref{coarse_top50}).
Personalized judges again dominate on most dimensions, with particularly large gains on technical validity and market size.

Together, these metrics confirm that the personalization advantage extends beyond fine-grained ordinal agreement to coarse selection behavior, especially on dimensions where expert disagreement is highest.

\begin{table}[t]
\centering
\small
\renewcommand{\arraystretch}{0.95}
\setlength{\tabcolsep}{4pt}
\begin{tabular}{lcc}
\toprule
Dim. & Aggregate & Personalized \\
\midrule
Spec. & 0.747 & 0.780 \\
Tech. val. & 0.594 & 0.714 \\
Innov. & 0.759 & 0.715 \\
Comp. adv. & 0.718 & 0.725 \\
Need & 0.483 & 0.574 \\
Market & 0.567 & 0.648 \\
\bottomrule
\end{tabular}
\caption{
Above-median Jaccard similarity between judge and expert selections, at 9-shot.
}
\label{tab:coarse_jaccard}
\end{table}

\subsection{Evaluator Agreement and Reasoning Structure}
\label{sec:reasoning_structure}

Beyond score alignment, we examine whether personalized judges also capture differences in evaluators' underlying reasoning patterns (\figref{reasoning_similarity}).
For evaluator pairs with overlapping idea–dimension annotations,
we compute two quantities.
First, we measure their empirical agreement using mean \krip{} across shared instances.
Second, we compute the cosine similarity between reasoning texts generated by the zero-shot, aggregate, and personalized judges.
Embeddings are generated by Qwen3-Embedding-8B~\citep{qwen3technicalreport}.
Under personalized conditioning, a positive relationship emerges between evaluator agreement and reasoning similarity (Pearson $r = 0.31$).
Evaluators who agree more on scores tend to produce more semantically similar reasoning when modeled individually.
In contrast, this relationship is near zero under aggregate or zero-shot configurations.

This result indicates that personalized judges do not merely replicate score distributions.
They capture evaluator-specific evaluation policies that manifest both in numeric scores and in textual justifications.
When two evaluators share similar standards, their personalized judges generate correspondingly similar reasoning.
When standards diverge, reasoning diverges as well.
These findings strengthen the interpretation that disagreement in \pbench{} reflects structured heterogeneity rather than noise.
Aggregate conditioning smooths differences,
while personalized conditioning preserves them.

\begin{table}[t]
\centering
\small
\renewcommand{\arraystretch}{0.95}
\setlength{\tabcolsep}{4pt}
\begin{tabular}{lcc}
\toprule
Dim. & Aggregate & Personalized \\
\midrule
Spec. & 0.594 & 0.638 \\
Tech. val. & 0.455 & 0.621 \\
Innov. & 0.681 & 0.560 \\
Comp. adv. & 0.660 & 0.638 \\
Need & 0.611 & 0.619 \\
Market & 0.486 & 0.667 \\
\bottomrule
\end{tabular}
\caption{
Top-50\% overlap: fraction of each expert's top-50\% items that the judge also ranks in its top 50\%, at 9-shot.
}
\label{tab:coarse_top50}
\end{table}

\section{Conclusion}
\label{sec:conclusion}

This paper examined business idea evaluation under heterogeneous expert standards.
Using \pbench{}, we showed that expert scores often fail to converge to a shared ordinal scale, even when a common rubric is provided.
Yet, agreement under coarse selection indicates that disagreement at the fine-grained level reflects calibrated but distinct evaluation policies rather than random noise.

We compared zero-shot, aggregate, and personalized judge configurations under these conditions.
Aggregate judges improve over rubric-only prompting but approximate an averaged consensus that does not fully represent any individual evaluator.
Personalized judges consistently achieve higher alignment with the corresponding evaluator and preserve evaluator-specific reasoning structure.
These findings suggest that, in pluralistic evaluation settings, the choice of target signal is not neutral: modeling a pooled standard and modeling individual standards lead to substantively different behaviors.

From an industry perspective, these results have practical implications for deploying LLM-based evaluation in ideation workflows.
In real-world organizations, business ideas are often assessed by stakeholders with different roles and incentives.
Collapsing their judgments into a single aggregate score may obscure meaningful differences in perspective.
Rather than enforcing artificial consensus, evaluation systems can surface structured disagreement and assist decision-makers in navigating it.

More broadly, our results indicate that evaluation for business ideation cannot be reduced to a single ground truth without losing information about stakeholder-specific judgment.
Future work should explore evaluation protocols that explicitly account for heterogeneous standards, including calibration mechanisms and multi-perspective scoring that preserve rather than suppress evaluator diversity.

\section*{Ethical Considerations and Licensing}

All data in \pbench{} are derived from public patent documents and automatically generated ideas submitted to an open shared task.
No personal or sensitive information is included.
Annotations for the computer science and NLP domains were performed by consenting professional
annotators and are released at \url{https://stockmarkteam.github.io/pbig-data/}. For the materials chemistry domain, annotations were conducted in collaboration with a private
company, and their release will follow contractual.

\bibliography{ref}


\clearpage
\appendix

\section{Full Scoring Rubric}
\label{sec:rubric_full}

This section provides the complete rubric used for expert scoring.
The rubric was designed to reflect practical business-oriented evaluation rather than purely linguistic quality.
Each dimension corresponds to a distinct aspect of product assessment that may be emphasized differently by different evaluators.

The six dimensions separate technical feasibility (specificity and technical validity), novelty and differentiation (innovativeness and competitive advantage), and commercial relevance (need validity and market size).
Need validity and market size are scored on a 0--3 scale in our rubric.

The rubric defines discrete score levels with written descriptions to encourage consistent interpretation.
However, as shown in the main text, even under a shared rubric, experts apply different implicit assumptions and thresholds.
We therefore release the full rubric to enable reproducibility and to allow future work to investigate alternative judge designs under the same evaluation specification.

\begin{table*}[t]
\centering
\small
\renewcommand{\arraystretch}{1.15}
\setlength{\tabcolsep}{3pt}
\begin{tabular}{m{2.2cm} m{3.8cm} m{9.6cm}}
\toprule
Criterion & Description & Rubric \\
\midrule

Specificity &
Clarity and concreteness of the product description. &
1. Cannot be read as coherent language;
2. Can be read as language, but the idea's meaning is barely conveyed;
3. One or more concrete products can be imagined;
4. A single concrete product can be clearly imagined.
\\ \midrule

Technical validity &
Feasibility of implementing the idea using the patent. &
1. The patented technology does not seem suitable for the use;
2. Building a prototype using the technology is challenging but possible;
3. A prototype could be built using the technology;
4. The technology can be applied to a production-level product.
\\ \midrule

Innovativeness &
Novelty and originality of the proposed solution. &
1. A well-known application; lacks novelty;
2. Known use case of similar technology, but not yet fully explored;
3. A use case I hadn't thought of, but not particularly exciting;
4. Surprising and novel; strong originality;
5. Clearly innovative and potentially groundbreaking.
\\ \midrule

Competitive advantage &
Distinct benefits and advantages over existing solutions. &
Two sub-questions are considered:
(A) Is it difficult to imitate the idea using the technology?
(B) Is the technology essential to the core of the idea?
\newline
1. Neither A nor B;
2. Only B;
3. Only A;
4. Both.
\\ \midrule

Need validity &
Relevance of the product to genuine user needs. &
0. Not a B2B product;
1. Both qualitative and quantitative returns are low;
2. Either quantitative (monetary) or qualitative (for corporate growth) returns are large;
3. Both qualitative and quantitative returns are large.
\\ \midrule

Market size &
Number of potential users. &
0. Not a B2B product;
1. Niche, appeals to some companies;
2. Many companies acknowledge the issue; adoption depends on budget/systems;
3. Necessary for almost all companies.
\\

\bottomrule
\end{tabular}
\caption{Full rubric for the six business-oriented scoring dimensions.}
\label{tab:rubric_full}
\end{table*}

\section{Overview of Idea Generation Systems}
\label{sec:gen_overview}

This section briefly summarizes the ideation systems whose outputs are included in \pbench{}.
The purpose of this overview is to clarify data provenance and diversity, not to compare generation performance.
The dataset includes ideas generated by multiple independent systems
\citep{yoshiyasu-2025-team,kanumolu-etal-2025-agent,
xu-etal-2025-mk2,terao-tachioka-2025-collaborative,
hoshino-etal-2025-business,shimanuki-etal-2025-business}.
These systems vary in prompting strategies, use of multi-agent interaction, and refinement mechanisms.
Some rely primarily on structured prompting,
while others employ multi-agent discussions, critique–revision loops,
or iterative rewriting pipelines.

Generation performance comparisons and system-level analyses
are reported in prior workshop publications
\citep{hirota-etal-2025-overview}.
In contrast, the present work does not evaluate or rank these systems.
Their outputs are used collectively to analyze evaluation behavior under heterogeneous expert standards.
Using ideas from multiple systems ensures that the observed disagreement patterns are not artifacts of a single generation strategy.
The conclusions of this paper therefore concern evaluation methodology rather than generation quality.

\section{Staged Screening Protocol}
\label{sec:screening_protocol}

\begin{figure}[t]
    \centering
    \includegraphics[width=\columnwidth]{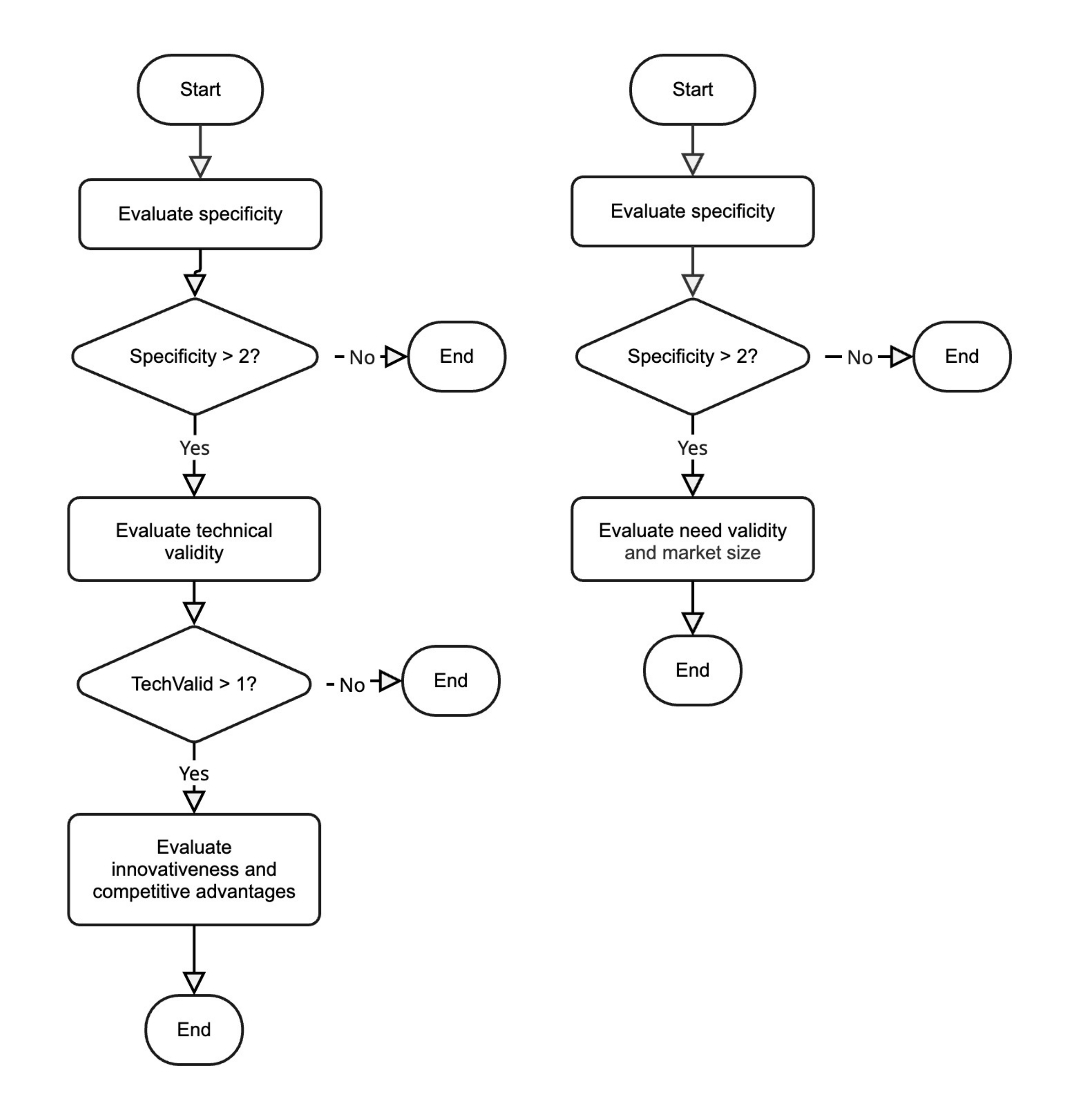}
    \caption{Staged screening protocol for expert scoring.}
    \label{fig:annotation_steps}
\end{figure}

\figref{annotation_steps} shows the staged screening protocol used during expert scoring.
Downstream dimensions are evaluated only when prerequisite conditions are met, which produces missing scores by design.

\section{Judge Prompt Schema}
\label{sec:prompt_schema}

\begin{figure*}[t]
\centering
\begin{tcolorbox}[title={\llmjudge{} Prompt Template}, breakable, width=0.95\textwidth]
\small
\begin{verbatim}
You are given a pair consisting of a patent and a product idea based on that
patent. Your task is to evaluate the idea following the given instruction.
First, you will receive a detailed instruction. If the setting is few-shot,
several examples of patents, ideas, and scores are also provided. Finally,
you will be given a new patent and idea to evaluate.

## Instruction
<Instruction text here>
## Examples (only in few-shot setting)
<Example 1: patent, idea, and corresponding score>
<Example 2: patent, idea, and corresponding score>
...
<Example N: patent, idea, and corresponding score>

## Input
<Patent and idea to be scored>
## Output format
Return a single line of valid JSON in this format:
{"score": <number>,
 "reason": "<brief reason>",
 "confidence": <integer between 0 and 100>}
\end{verbatim}
\end{tcolorbox}
\caption{The prompt template for \llmjudge{} models.
Angle brackets (<...>) denote placeholders.}
\label{fig:prompt_template_simplified}
\end{figure*}

All judge configurations use the same prompt structure, shown in \figref{prompt_template_simplified}.

\section{Confidence Filtering Discard Rates}
\label{sec:discard_rates}

Predictions whose self-reported confidence value is below 80 are discarded, as described in Section~\ref{sec:judges}.
\tabref{discard_rates} reports the proportion of discarded predictions per dimension and domain.
All discard rates are below 0.32\%, so the filtering step has a negligible effect on the reported alignment patterns.

\begin{table}[t]
\centering
\small
\renewcommand{\arraystretch}{1}
\setlength{\tabcolsep}{4pt}
\begin{tabular}{lrrr}
\toprule
Dim. & NLP & CS & Mat. \\
\midrule
Spec. & 0.012\% & 0.103\% & 0.319\% \\
Tech. val. & 0.026\% & 0.008\% & 0.108\% \\
Innov. & 0.000\% & 0.000\% & 0.000\% \\
Comp. adv. & 0.001\% & 0.003\% & 0.006\% \\
Need & 0.004\% & 0.000\% & 0.003\% \\
Market & 0.000\% & 0.000\% & 0.000\% \\
\bottomrule
\end{tabular}
\caption{
Proportion of predictions discarded by confidence filtering (confidence $<$ 80), by dimension and domain.
}
\label{tab:discard_rates}
\end{table}

\section{Qualitative Case Study}
\label{sec:case_study}

To illustrate how aggregate and personalized judges can diverge even for the same item, we examine an innovativeness assessment of an idea that applies a patented UI-customization technology to electronic medical record (EMR) systems with role-based permissions.
The target evaluator assigned a conservative innovativeness score of 2, reflecting the view that role-based EMR customization, while a plausible extension, is a known use case within healthcare IT.
The personalized judge reproduced this score and justified it by noting that ``\textit{the idea applies the patented UI customization technology to EMR systems with role-based permissions, which is a known use case in healthcare IT, but the specific implementation for dynamic clinical workflows is not yet widely explored}.''
The aggregate judge, conditioned on mixed-evaluator histories, instead returned a score of 4 and framed the same idea more generously as ``\textit{a plausible but not obvious extension of the patent \ldots\ beyond generic customization by integrating into clinical workflows and compliance needs}.''
The two judges thus agree on the qualitative description of the idea but apply different innovation thresholds.
Under personalized conditioning, the judge inherits the target evaluator's conservative calibration; under aggregate conditioning, the mixed history pulls the score toward a more permissive pooled standard that no individual evaluator necessarily holds.

\section{Results for GPT-5 mini}
\label{sec:gpt5_results}

To test whether the personalization advantage generalizes beyond the Qwen3 family, we repeat the main judge comparison with gpt-5-mini-2025-08-07.
\tabref{gpt5_alpha} reports Krippendorff's $\alpha$ between the judge predictions and expert annotations for the aggregate and personalized configurations.
Personalized conditioning yields alignment that is higher than or comparable to the aggregate judge on five of the six dimensions, with the largest gaps on need validity and market size.
This mirrors the pattern observed for Qwen3 models in Section~\ref{sec:judges} and supports the claim that the personalization advantage is not an artifact of a single model family.

\begin{table}[t]
\centering
\small
\renewcommand{\arraystretch}{0.95}
\setlength{\tabcolsep}{4pt}
\begin{tabular}{lrr}
\toprule
Dim. & Aggregate & Personalized \\
\midrule
Spec. & $-0.083$ & $0.084$ \\
Tech. val. & $-0.042$ & $-0.012$ \\
Innov. & $0.235$ & $0.233$ \\
Comp. adv. & $-0.028$ & $-0.119$ \\
Need & $0.189$ & $0.404$ \\
Market & $0.048$ & $0.257$ \\
\bottomrule
\end{tabular}
\caption{
Krippendorff's $\alpha$ between gpt-5-mini-2025-08-07 judge predictions and expert annotations, under aggregate and personalized conditioning.
}
\label{tab:gpt5_alpha}
\end{table}

\end{document}